\documentclass[12pt]{article}

\usepackage[preprint, nonatbib]{neurips_2023}

\usepackage[utf8]{inputenc} 
\usepackage[T1]{fontenc}    
\usepackage{hyperref}       
\usepackage{url}            
\usepackage{booktabs}       
\usepackage{amsfonts}       
\usepackage{nicefrac}       
\usepackage{microtype}      
\usepackage{xcolor}         

\usepackage{fancyhdr}

\usepackage{algorithm}
\usepackage{algpseudocode}

\hypersetup{colorlinks, linkcolor=blue, citecolor=blue, urlcolor=blue}

\usepackage{amsfonts,dsfont}
\usepackage{hyperref}
\usepackage{color}
\usepackage{amsmath, amssymb}
\usepackage{amsthm}
\usepackage{bm}
\usepackage{stmaryrd}
\usepackage{color}
\usepackage{comment}
\usepackage{geometry}
\allowdisplaybreaks

\usepackage{graphicx}
\usepackage{caption}
\usepackage{subcaption}

\usepackage{tabularx}

\usepackage{caption}

\newcommand{\todo}[1][\null]{\ensuremath{\clubsuit}}

\newcommand{\noprint}[1]{}

\theoremstyle{definition}\newtheorem{definition}{Definition}
\theoremstyle{definition}
\theoremstyle{definition}\newtheorem{remark}[definition]{Remark}
\theoremstyle{definition}
\theoremstyle{definition}
\theoremstyle{definition}
\theoremstyle{definition}
\theoremstyle{definition}
\theoremstyle{definition}

\title{Exactly conservative physics-informed neural networks and deep operator networks\\ for dynamical systems}

\author{%
  Elsa Cardoso-Bihlo \\
  Department of Mathematics and Statistics\\
  Memorial University of Newfoundland\\
  St. John’s, NL, A1C 5S7, Canada \\
  \texttt{ecardosobihl@mun.ca} \\
  \And
  Alex Bihlo \\
  Department of Mathematics and Statistics\\
  Memorial University of Newfoundland\\
  St. John’s, NL, A1C 5S7, Canada \\
  \texttt{abihlo@mun.ca} \\
}

\begin{document}

\maketitle

\vspace{0.25cm}

\begin{abstract}
We introduce a method for training exactly conservative physics-informed neural networks and physics-informed deep operator networks for dynamical systems. The method employs a projection-based technique that maps a candidate solution learned by the neural network solver for any given dynamical system possessing at least one first integral onto an invariant manifold. We illustrate that exactly conservative physics-informed neural network solvers and physics-informed deep operator networks for dynamical systems vastly outperform their non-conservative counterparts for several real-world problems from the mathematical sciences.
\end{abstract}

\section{Introduction}

Recent years have seen a renewed interest in solving differential equations using neural networks. The seminal contribution on this subject dates back to the work of~\cite{laga98a}, and is now referred to as \textit{physics-informed neural networks} (PINNs) \cite{rais19a} when approximating the solution of a differential equation directly and \textit{physics-informed deep operator networks} (DeepONets) \cite{lu21a} when approximating the solution operator instead, respectively. This interest in using neural networks for solving differential equation parallels the breakthroughs of modern machine learning in general~\cite{brow20a,kriz12a,silv17a}, and has given rise to the emerging discipline of \textit{scientific machine learning}~\cite{cuom22a}.

While physics-informed neural networks and DeepONets are seeing a strong surge in their popularity, they exhibit several shortcomings that have led to the development of numerous improvements over the past several years, see the following Section~\ref{sec:RelatedWork} for a more in-depth review. In our paper we address one of these shortcomings, that has not received adequate attention in the literature yet: \textit{The lack of inherent preservation of conservation laws in these networks.} Both physics-informed neural networks and physics-informed DeepONets are trained to satisfy a given system of differential equations over a specific (spatio-)temporal domain, but the exact preservation of \textit{first integrals} (for the case of dynamical systems) or \textit{conservation laws} (for the case of partial differential equations) is not automatically guaranteed by their training procedure. Indeed, as we will explicitly show in this paper, standard physics-informed neural networks and DeepONets exhibit a rather strong drift in first integrals for dynamical systems over time, which also leads to spurious artifacts in their numerical solutions as well. We then propose a remedy for this violation of the important first principle of conservation of first integrals by introducting exactly conservative physics-informed neural networks and physics-informed DeepONets.

The further organization of this paper is given as follows. Section~\ref{sec:RelatedWork} provides a short summary on the related relevant work within the areas of geometric numerical integration and physics-informed neural networks. In Section~\ref{sec:Method} we present our method for the construction of exactly conservative neural network based solvers for dynamical systems. The numerical examples illustrating this method are collected in Section~\ref{sec:Examples}. These examples contain both conservative physics-informed neural networks and conservative physics-informed DeepONets. The final Section~\ref{sec:Conclusion} summarizes the findings of this paper and presents some concluding thoughts about potential future research avenues.

\section{Related work}\label{sec:RelatedWork}

Physics-informed neural networks were introduced in~\cite{laga98a} and more recently popularized in~\cite{rais19a}. The idea of this method is to approximate the solution of a system of differential equations through a global solution interpolant parameterized by a neural network. The neural network is trained to satisfy the differential equation along with any relevant initial and/or boundary conditions over a collection of collocation points. Once trained, the neural network ideally becomes a numerical approximation of the exact solution of the given differential equation.

Numerous issues of physics-informed neural networks have been identified, such as their inability to converge to the right solution of many differential equations that are solved over larger time-domains, along with certain mitigation strategies~\cite{bihl23a, bihl22a, wang23a, wang22a}. Physics-informed neural networks are also costly to train, and may be many orders of magnitude slower than standard numerical methods today~\cite{brec23b}.

Despite the increasing popularity of neural network based differential equations solver, an issue that has received considerably less attention is their ability to preserve \textit{geometric properties} of differential equations. The study of geometric properties of numerical approximations to differential equations is a prolific research field within the numerical analysis of differential equations, where it is referred to as \textit{geometric numerical integration}. Geometric numerical integration is devoted to the development of numerical schemes preserving a wide range of qualitative properties of differential equations, such as symplectic structures~\cite{sanz94a}, Lie symmetries~\cite{bihl17b}, and first integrals~\cite{wan16a}. For a more in-depth discussion of geometric numerical integration, we refer to the standard textbook~\cite{hair06Ay}.

To the best of our knowledge, the analogous field of \textit{geometric numerical machine learning} is still in its infancy, and very little explicit consideration of geometric properties of the solutions obtained using physics-informed neural networks is available in the literature. Invariant physics-informed neural networks were recently proposed in~\cite{aror23a}. ``Conservative'' physics-informed neural networks were introduced in~\cite{jagt20a}, which rely on domain decomposition approaches and the use of Gauss' theorem to improve upon standard physics-informed neural networks. Importantly though, these networks are not conservative in the sense of geometric numerical integration, as they do not preserve conservation laws up to machine precision. Most closely related to our paper, exact preservation of first integrals in neural networks for dynamical systems was recently considered in~\cite{mulle23a}, using Noether's theorem to enforce Lie symmetries in neural networks, which for Lagrangian systems then automatically implies exact conservation of the associated first integrals. A main restriction of this approach is that it only works for Lagrangian systems, as non-Lagrangian systems are outside of the scope of Noether's theorem. We should also like to refer to the growing body of work on equivariant neural networks in this regard, see e.g.~\cite{finz21a}. For Lagrangian dynamical systems, equivariant neural network are again intimately linked to the preservation of first integrals.

\section{Method}\label{sec:Method}

In this section we introduce a method for exactly conserving first integrals for arbitrary dynamical systems. We do this by first recalling some background information on first integrals of dynamical systems and physics-informed neural networks.

\subsection{First integrals of dynamical systems}

We collect some relevant notions on first integrals of systems of ordinary differential equations here. 

Consider the first order system of ordinary differential equations
\begin{equation}\label{eq:GeneralSystem}
  \frac{\mathrm{d}u}{\mathrm{d}t} = f(t,u),\quad u(t_0) = u_0.
\end{equation}
where $u\in\mathbb{R}^n$ and $f\colon\mathbb{R}\times\mathbb{R}^n\to\mathbb{R}^n$, where we assume sufficient regularity of $f$ such that a unique solution $u=u(t)$ exists for all $t\in\mathbb{R}$. A first integral of system~\eqref{eq:GeneralSystem} is a continuously differentiable function $I\colon\mathbb{R}\times\mathbb{R}^n\to\mathbb{R}$ such that
\[
\frac{\mathrm{d}I}{\mathrm{d}t}=0
\]
holds for any solution of system~\eqref{eq:GeneralSystem}. In other words, $I=I(t,u)$ satisfies the equation
\[
\frac{\partial I}{\partial t} + \frac{\partial I}{\partial u}\frac{\mathrm{d}u}{\mathrm{d}t} = \frac{\partial I}{\partial t} + \frac{\partial I}{\partial u}f(t,u) = 0.
\]
This implies that $I(t,u) = I(t_0,u_0)=I_0$ for all solutions of system~\eqref{eq:GeneralSystem}, i.e.\ the value of $I$ remains constant along solutions of system~\eqref{eq:GeneralSystem}. For the first integral to be non-trivial, we assume that $I$ is a non-constant function of its arguments. 

To simplify the notation for what follows, we assume without loss of generality that the first integrals do not explicitly depend on time, i.e.\ $I=I(u)$\footnote{The time-dependent case can be considered analogously to how non-autonomous systems of differential equations can be converted to autonomous systems by introducing $t$ as another dependent variable in the system~\eqref{eq:GeneralSystem}.}. Subsequently, we refer to the $n-m$-dimensional sub-manifold of $\mathbb{R}^n$ spanned by the intersection of the iso-surfaces of $m\leqslant n$ functionally independent first integrals $I^i$, $i=1,\dots,m$, of system~\eqref{eq:GeneralSystem} as $\mathcal M$, i.e.\ 
\[
\mathcal M = \{u\in \mathbb{R}^{n-m} | I^1(u)-I^1_0=0,\dots, I^m(u)-I^m_0=0)\}=\{u\in \mathbb{R}^{n-m} | g(u)=0\}.
\]

\subsection{Physics-informed neural networks and physics-informed DeepONets}

Physics-informed neural networks for solving system~\eqref{eq:GeneralSystem} were originally proposed in~\cite{laga98a}. The idea is to seek the solution to system~\eqref{eq:GeneralSystem} parameterized as a neural network of the form $u^{\boldsymbol{\theta}}=\mathcal N^{\boldsymbol{\theta}}(t)$, where $\boldsymbol{\theta}$ is the vector of weights of the underlying neural network $\mathcal N$. This network is trained to satisfy the given dynamical system over the fixed temporal domain $[t_0,t_{\rm f}]$ by sampling a collection of $N$ collocation points $\{t_i\}_{i=0}^N$ such that $t_0=t_0<t_1<\cdots<t_N=t_{\rm f}$, which in practice is done by minimizing the following physics-informed loss function:
\begin{equation}\label{eq:PINNloss}
    \mathcal{L}(\boldsymbol{\theta})= \sum_{i=0}^N\left(\frac{\mathrm{d}u^{\boldsymbol{\theta}}}{\mathrm{d}t}\bigg|_{t=t_i}-f(t_i,u^{\boldsymbol{\theta}}(t_i))\right).
\end{equation}
The derivative in the above loss function is then typically computed using automatic differentiation, the same algorithm that is at the heart of backpropagation, which is used to train neural networks and thus readily available in the most prominent deep learning toolkits such as \texttt{JAX}, \texttt{PyTorch} and \texttt{TensorFlow}.

To guarantee that the trained neural network approximation to the exact solution of system~\eqref{eq:GeneralSystem} also satisfies the initial condition, we structure the neural network solution as
\[
u^{\boldsymbol{\theta}}(t) = \mathcal N^{\boldsymbol{\theta}}(t) = u_0 + \frac{t-t_0}{t_{\rm f}-t_0}\mathcal{\tilde N}^{\boldsymbol{\theta}}(t),
\]
where $\mathcal{\tilde N}^{\boldsymbol{\theta}}$ is a standard multi-layer perceptron. This ansatz guarantees that even before training $u^{\boldsymbol{\theta}}(t_0)=u_0$ and thus the initial condition is automatically exactly satisfied. More complicated ansatzes could be use to enforce the initial condition, cf.~\cite{brec23c}, but for the problems studied below we found that the above ansatz works sufficiently well.

\begin{remark}
Most implementations of the above described algorithms for training physics-informed neural networks in the literature do not include the initial condition as a hard constraint in the neural network. In this case, the physics-informed loss function would require a second component that enforces the initial condition as a soft constraint on the neural network~\cite{rais18a}. For the consideration of exactly conservative neural networks, it is critical that the true initial condition is exactly satisfied by the neural network solution, since this initial condition determines the correct value of the first integrals over the entire interval of integration.
\end{remark}

One of the shortcomings of physics-informed neural networks is that they are challenging to train for long time intervals~\cite{bihl22a,kris21a,wang23a}. It is generally found that if $t_{\rm f}\gg t_0$ then physics-informed neural networks converge to a trivial solution of the underlying system of differential equations. To remedy this situation, it is either possible to split the integration interval into sub-intervals and train individual physics-informed neural networks for each sub-interval~\cite{bihl22a,kris21a}, or to train physics-informed deep operator networks instead~\cite{lu21a,wang23a}. 

Physics-informed deep operator networks (DeepONets) are conceptually similar to physics-informed neural networks, except that they try to learn the solution operator of a differential equation instead of a particular solution itself. That is, if $\mathfrak{G}(u_0)$ is the corresponding solution operator for system~\eqref{eq:GeneralSystem} such that $u(t)=\mathfrak{G}(u_0)(t)$ is the exact solution for all $t\in\mathbb{R}$, the goal of DeepONets is to approximate $\mathcal G^{\boldsymbol{\theta}}(u_0)(t)\approx \mathfrak{G}(u_0)(t)$ using a neural network. This neural network typically consists of two parts, a branch net with input $u_0$ and a trunk net with input $t$, where both branch and trunk networks are generally standard multi-layer perceptrons. The outputs of these networks are finally combined using a scalar product, to produce the solution $u^{\boldsymbol{\theta}}(t)=\mathcal G^{\boldsymbol{\theta}}(u_0)(t)$, see Fig.~\ref{fig:cDeepONet} (blue) for a schematic overview. The theoretical underpinning of DeepONets is the universal approximation theorem for operators, first formulated in~\cite{chen95a}. Once a DeepONet has been trained for a fixed time interval $[t_0,\Delta t]$ with $\Delta t\ll t_{\rm f}$, it can be used to obtain the solution for arbitrary long time intervals $[t_0,t_{\rm f}]$ using simple time stepping~\cite{wang23a}.

\subsection{Conservative neural network based solvers using projection methods}

We first recall the standard project method from geometric numerical integration~\cite{hair06Ay}. Let $\Phi_{\Delta t}$ be an arbitrary one-step numerical method with fixed time step $\Delta t$, such that for a given $u_n$, approximating the exact solution $u(t_n)$ of system~\eqref{eq:GeneralSystem} at time step $t_n=t_0+n\Delta t$, $t_n\in[t_0,t_\mathrm{f}-\Delta t]$, we have $\tilde u_{n+1} = \Phi_{\Delta t}(t_n,u_n)$. 

For general one-step methods, there is no guarantee that $\tilde u_{n+1}\in\mathcal M$, even if $u_n\in\mathcal M$. Thus, after each step with the one-step method, the solution $\tilde u_{n+1}$ is projected onto $M$ by solving a constrained minimization problem, namely
\begin{equation}\label{eq:ProjectionConstrained}
 ||u_{n+1} - \tilde u_{n+1} || \to \mathrm{min}\qquad \mathrm{subject\ to}\qquad g(u_{n+1})=0.
\end{equation}
Once successfully solved, the projected solution indeed satisfies $u_{n+1}\in\mathcal M$, and the process is repeated after each step until the final integration time is reached. In practice, the constrained minimization problem can be solved using a few steps of Newton iteration, see~\cite{hair06Ay} for further details.

The key assumption of any projection method is that the candidate solution is close to the manifold where the constrained solution should live. Unfortunately, for general conserved surfaces consisting of several connected components there is no guarantee that the projection method will project the candidate solution onto the right connected component~\cite{wan16a}. In other words, even for increasingly small step sizes, the projection step may fail.

We note here that the constrained optimization problem~\eqref{eq:ProjectionConstrained} does not depend on $\tilde u_{n+1}$ being computed with a one-step method. Indeed, let $\{\tilde u_i^{\boldsymbol{\theta}}\}_{i=1}^N$ be the numerical solution obtained using a standard physics-informed neural network or DeepONet at the collocation points $\{t_i\}_{i=1}^N$, this solution can also be projected onto the invariant manifold $\mathcal M$ by solving the same constrained optimization problem~\eqref{eq:ProjectionConstrained} at all collocation points. 

\begin{figure}[!ht]
\centering
\begin{subfigure}[b]{0.45\textwidth}
\includegraphics[width=\textwidth]{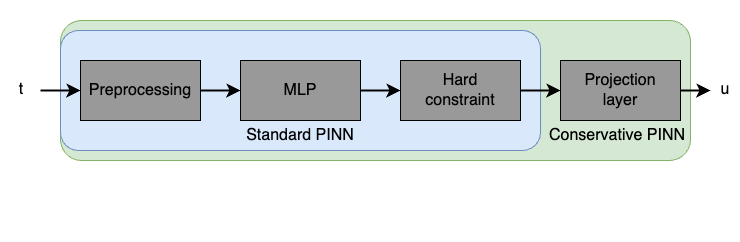}
\caption{Conservative PINNs.}
\end{subfigure}
\begin{subfigure}[b]{0.45\textwidth}
\includegraphics[width=\textwidth]{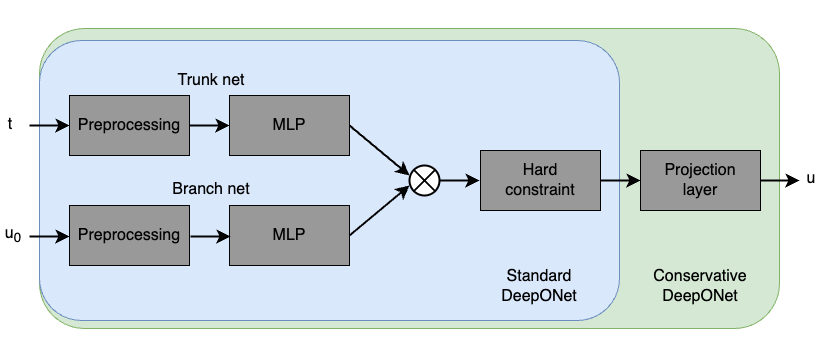}
\caption{Conservative DeepONets.}
\label{fig:cDeepONet}
\end{subfigure}
\caption{Architectures of standard vs.\ conservative physics-informed neural networks and physics-informed DeepONets.}
\label{fig:cPINN}
\end{figure}

We thus propose the following extension to hard constrained physics-informed neural networks and physics-informed DeepONets: We add a conservative projection layer as the last layer to the network that solves the constrained optimization problem~\eqref{eq:ProjectionConstrained} using a simplified Newton iteration, cf.~Fig.~\ref{fig:cPINN}. We should like to note here that this conservative projection layer is compatible with the hard constrained layer, which is the penultimate layer in the network. This is because at $t=t_0$ the hard constrained layer will yield the true initial condition which by definition also lies on the manifold $\mathcal M$. We denote with $u^{\boldsymbol{\theta}_e}_{\rm st}(t)=\mathcal N^{\boldsymbol{\theta}_e}(t)$ the standard physics-informed neural network solution after $e\in\{0,\dots,E\}$ epochs of training, and with $u^{\boldsymbol{\theta}_e,p}_{\rm c}(t) = \mathsf{P}_p\mathcal N^{\boldsymbol{\theta}_e}(t)=\mathsf{P}_pu^{\boldsymbol{\theta}_e}_{\rm st}(t)$ the associated conservative physics-informed neural network using a total of $p\in\{0,\dots,P\}$ Newton iterations to solve the constrained optimization problem~\eqref{eq:ProjectionConstrained}. Thus, the projection step could also be applied to already trained (non-conservative) physics-informed neural network, with the caveat that the solution would have to be sufficiently close to $\mathcal M$ so that the projection does not fail.

For training conservative neural network solvers from scratch, we note that the projection method may fail as the initial estimate for the numerical solution in the early stages of training will be wrong for $t\gg t_0$, and thus potentially far away from the true connected component of $\mathcal M$ onto which the solution would have to be projected. To account for this we, we introduce a projection schedule by setting $p=p(e)$, thus making the number of projection steps a function of the training epochs, such that $p(0)=0$ and $p(E)=P$. In other words, we begin to train conservative physics-informed neural networks as standard physics-informed neural networks, and as training progresses, we increase the number of projection steps $p$ at fixed intervals, thus mapping the solution $u^{\boldsymbol{\theta}_e,p}_{\rm c}$ more and more onto $\mathcal M$. We choose to increase $p$ 
as a function of the training epoch by setting $p(e)=\lfloor e(P+1)/E\rfloor$, where $\lfloor\cdot\rfloor$ is the floor function, so that $p$ increases as a step function in equal intervals over training, although different projection scheduler would be possible as well. In addition, to prevent abrupt jumps in the learned solution $u^{\boldsymbol{\theta}_e,p}_{\rm c}$ as $p$ is increased in integer steps over training, we use a soft update for $u^{\boldsymbol{\theta}_e,p}_{\rm c}$ in the form of
\[
u^{\boldsymbol{\theta}_e,p}_{\rm c}(t) \leftarrow (1-s(e))u^{\boldsymbol{\theta}_e,p-1}_{\rm c}(t)+s(e)u^{\boldsymbol{\theta}_e,p}_{\rm c}(t),\quad p \geqslant 1,
\]
where $s(e)$ is a weight function such that $s(0)=0$ and $s(E)=1$. For the sake of simplicity, in the following we used $s(e)=e/E$. The entire training algorithm for conservative physics-informed neural networks is summarized in Algorithm~\ref{alg:ConservativePINNs}.

\begin{algorithm}[!ht]
\caption{Exactly conservative PINNs for solving~\eqref{eq:GeneralSystem} over the interval $[t_0,t_{\rm f}]$.}\label{alg:ConservativePINNs}
\begin{algorithmic}
\Require A neural network $u^{\boldsymbol{\theta}_0,p}(t)=\mathsf{P}_p\mathcal N^{\boldsymbol{\theta}_0}(t)$ satisfying $u^{\boldsymbol{\theta}_0,p}(t_0)=u_0$.
\Require A set of collocation points $\{t_i\}_{i=0}^N\in[t_0, t_{\rm f}]$. 
\Require Learning rate $\eta$, number of epochs $E\geqslant0$, and number of projection steps $P \geqslant 0$.
\Require A weight function $s(e)$ such that $s(0)=0$ and $s(E)=1$.
\Require A projection schedule $p(e)$ such that $p(0)=0$ and $p(E)=P$.
\While{$e\neq E$}
\State $e \gets e+1$
\State Update weight function $s(e)$ and projection schedule $p(e)$.
\State Compute $u^{\boldsymbol{\theta},p}=\mathsf{P}_p\mathcal N^{\boldsymbol{\theta}_e}(t)$.
\If{$p>0$}
\State Compute $u^{\boldsymbol{\theta}_e,p-1}=\mathsf{P}_{p-1}\mathcal N^{\boldsymbol{\theta}_e}(t)$
\State Use soft update $u^{\boldsymbol{\theta}_e,p}(t)\gets (1-s(e))u^{\boldsymbol{\theta}_e,p-1}(t)+s(e)u^{\boldsymbol{\theta}_e,p}(t)$
\EndIf
\State Compute gradient $\nabla\mathcal L$ of loss function~\eqref{eq:PINNloss}.
\State Update weights $\boldsymbol{\theta}_e \gets \boldsymbol{\theta}_e - \eta \nabla\mathcal L$.
\EndWhile
\end{algorithmic}
\end{algorithm}

Note that the corresponding algorithm for conservative physics-informed DeepONets is analogous to the one presented above for conservative physics-informed neural networks and is thus not shown here. For simplicity we indicated the weight update step here using standard gradient descent, but in practice we use Adam in the following.

\subsection{Implementation and training}

We implemented the above method using \texttt{JAX} using double precision arithmetic. All neural networks use a total of 4 hidden layers with 40 units per layer (for the DeepONets this is 4 hidden layers and 40 units per layer for each the branch and the trunk nets), with the hyperbolic tangent as activation function. We use Adam as our optimizer, with a learning rate of $\eta=10^{-3}$, and train all neural networks for a total of 5000 epochs using batch-gradient descent. All physics-informed neural networks trained below use a total of 2000 collocation points. The DeepONets trained use a total of 5000 collocation points (and thus initial conditions). For the conservative neural networks, we typically select the number of projection steps $P$ large enough so that the resulting numerical solution is conservative up to double precision machine epsilon. 

The notebooks for reproducing the numerical results reported below will be made available on Github upon publication of the article\footnote{\url{https://github.com/abihlo/ExactlyConservativeNeuralNetworks}}.

\section{Examples}\label{sec:Examples}

In this section we present several examples for standard and conservative physics-informed neural networks and DeepONets. For reference, we also present a numerical solution obtained using a standard fourth order Runge--Kutta integrator. This integrator is not conservative (except for linear conserved quantities) and should only provide a baseline comparison for the solution obtained using the two physics-informed neural networks, as it is one of the default numerical integrators in numerical analysis. Indeed, if a general dynamical system is being solved, Runge--Kutta methods are typically the default choice for the numerical integrator. We do not intent to draw conclusions from the comparison between the numerical results obtained using the Runge--Kutta integrator and the physics-informed neural networks, as it is difficult to provide a proper fair and quantitative comparison between machine-learning based and standard numerical methods in terms of approximation order, convergence properties and computational costs, and such a comparison is outside of the scope of this article. We should also like to mention that exactly conservative numerical schemes using classical geometric numerical integrators can be derived for all of the following examples~\cite{hair06Ay,wan16a}.

\subsection{Harmonic oscillator}

The harmonic oscillator is a classical mechanical system modeling the motion of a point particle with mass $m$ attached to a massless spring with spring constant $k$. The Hamiltonian for this problem is the total energy of the system
\[
H(q, p) = \frac{p^2}{2m} + \frac12kq^2,
\]
where $(q,p)$ is the position--momentum pair of the point particle moving through phase space. The (canonical Hamiltonian) equations of motion are then
\[
\frac{\mathrm{d}q}{\mathrm{d}t} = \frac{\partial H}{\partial p} = \frac{p}{m},\qquad \frac{\mathrm{d}p}{\mathrm{d}t} = -\frac{\partial H}{\partial q} = -kq,
\]
and the Hamiltonian remains conserved, $H(q(t), p(t))=H(q(0), p(0))$ for all $t\in\mathbb{R}$~\cite{gold80Ay}.

\begin{figure}[!ht]
\centering
\begin{subfigure}[b]{0.45\textwidth}
\includegraphics[width=\textwidth]{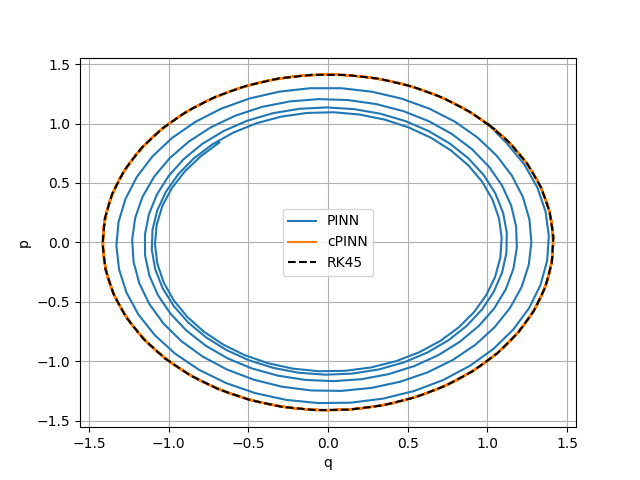}
\caption{Solution in the $(q,q)$ plane.}
\end{subfigure}
\begin{subfigure}[b]{0.45\textwidth}
\includegraphics[width=\textwidth]{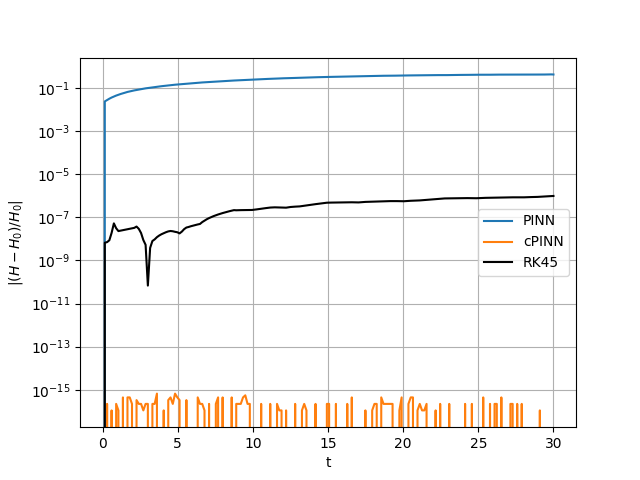}
\caption{Conservation of the Hamiltonian.}
\end{subfigure}
\caption{Numerical results for the harmonic oscillator.}
\label{fig:HarmonicOscillatorNumericalResults}
\end{figure}

We train vanilla and conservative physics-informed neural networks to solve the harmonic oscillator. The initial conditions are $u_0=(1,1)$ and the numerical solution is sought over the interval $t\in[0,30]$, for a mass of $m=1$ and a spring constant of $k=1$. The results obtained from the trained neural networks are depicted in Figure~\ref{fig:HarmonicOscillatorNumericalResults}.

These results show that standard physics-informed neural networks do not conserve the total energy. Indeed, energy is dissipated in this network, leading to a spurious drift of the solution equivalent to a damping effect. In turn, the conservative physics-informed neural network preserves the energy of the harmonic oscillator up to machine precision, and is thus not incurring any spurious damping effect, with the solution in phase space being a closed orbit.

\subsection{Rigid body}

The Euler equations for the free rigid body in angular momentum representation are
\begin{align}
\begin{split}
   &\frac{\mathrm{d}w_1}{\mathrm{d}t} = \frac{I_2-I_3}{I_2I_3}w_2w_3,\\
   &\frac{\mathrm{d}w_2}{\mathrm{d}t} = \frac{I_3-I_1}{I_1I_3}w_1w_3,\\
   &\frac{\mathrm{d}w_3}{\mathrm{d}t} = \frac{I_1-I_2}{I_1I_2}w_1w_2,
\end{split}
\end{align}
where $w=(w_1,w_2,w_3)$ is the angular momentum in the rigid body frame with $I_1$, $I_2$, $I_3$ being the principal moments of inertia~\cite{hair06Ay,holm09a}. The rigid body equations have the following two first integrals,
\[
H = \frac12\left(\frac{w_1^2}{I_1}+\frac{w_2^2}{I_2}+\frac{w_3^2}{I_3}\right),\quad L=\frac12(w_1^2+w_2^2+w_3^3),
\]
signifying kinetic energy and square of angular momentum preservation, respectively. The rigid body is also an example for a Nambu system~\cite{namb73Ay}.

\begin{figure}[!ht]
\centering
\begin{subfigure}[b]{0.45\textwidth}
\includegraphics[width=\textwidth]{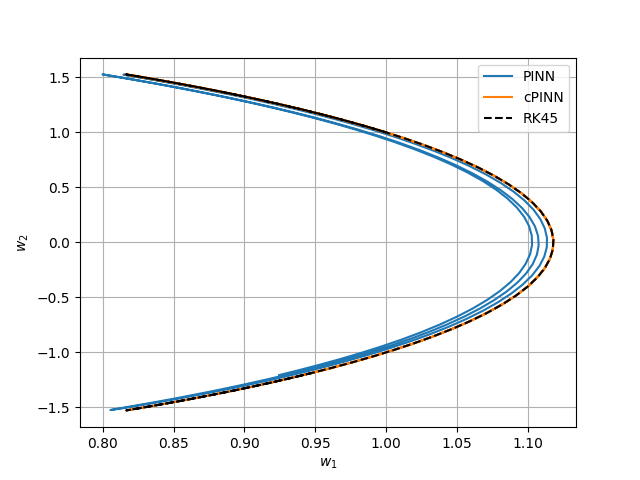}
\caption{Solution in the $(w_1,w_2)$ plane.}
\end{subfigure}
\begin{subfigure}[b]{0.45\textwidth}
\includegraphics[width=\textwidth]{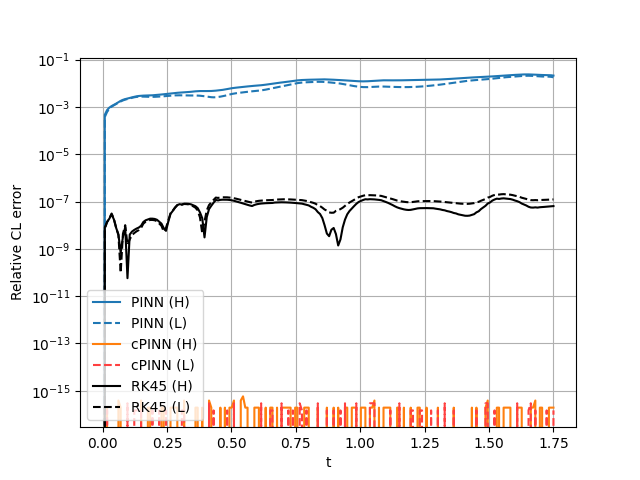}
\caption{Conservation of the first integrals.}
\end{subfigure}
\caption{Numerical results for the rigid body.}
\label{fig:RigidBodyNumericalResults}
\end{figure}

We train both standard and conservative physics-informed neural networks for the rigid body. We set $I_1=0.1$, $I_2=0.2$ and $I_3=0.3$ and train the networks to obtain the solution with initial condition $u_0=(1,1,1)$ for the time interval $t\in[0,1.75]$. The results of the numerical integration are presented in Figure~\ref{fig:RigidBodyNumericalResults}.

The motion of the rigid body should evolve along the intersection of the isosurfaces of the kinetic energy and square of angular momentum defined by the implicit equations $H-H_0=0$ and $L-L_0=0$, and should thus be closed and periodic. The violation of conservation of these first integrals in the standard physics-informed neural networks leads to the orbits of the solution not being closed, whereas the conservative physics-informed neural networks indeed move along the true intersection of the conserved surfaces. Both $H$ and $L$ are indeed exactly preserved by the conservative physics-informed neural network.

\subsection{Double pendulum}

The double pendulum is a canonical Hamiltonian system of two masses joined by two weightless rigid rods, one connecting the first mass to the origin, and one connecting the two masses~\cite{levi93a}. The motion is constrained in a plane, and the evolution of the system is determined by the respective angles $q_1$ and $q_2$ of the two rods, as well as the momenta of the two point masses $p_1$ and $p_2$. 

The Hamiltonian function for this system reads
\[
H = \frac{m_2l_2^2p_1^2+(m_1+m_2)l_1^2p_2^2 - 2m_2l_1l_2p_1p_2\cos(q_1-q_2)}{2m_2l_1^2l_2^2(m_1+m_2\sin^2(q_1-q_2)} - (m_1+m_2)gl_1\cos q_1 - m_2gl_2\cos q_2,
\]
where $m_1$ and $m_2$ are the masses of the two pendulums, $l_1$ and $l_2$ are the lengths of the two rods and $g$ is the gravitational constant. The associated equations of motion then follow from the canonical Hamiltonian equations
\[
\frac{\mathrm{d}q_i}{\mathrm{d}t} = \frac{\partial H}{\partial p_i},\quad \frac{\mathrm{d}p_i}{\mathrm{d}t} = -\frac{\partial H}{\partial q_i},\quad i=1,2.
\]
The double pendulum is a chaotic dynamical system~\cite{levi93a}.

\begin{figure}[!ht]
\centering
\begin{subfigure}[b]{0.45\textwidth}
\includegraphics[width=\textwidth]{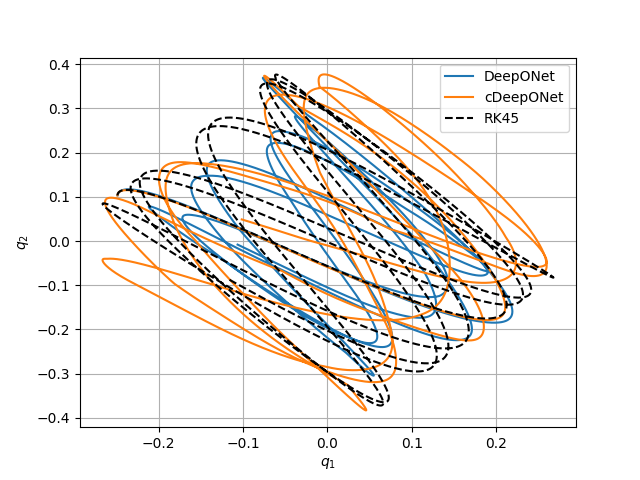}
\caption{Solution in the $(q_1,q_2)$ plane.}
\end{subfigure}
\begin{subfigure}[b]{0.45\textwidth}
\includegraphics[width=\textwidth]{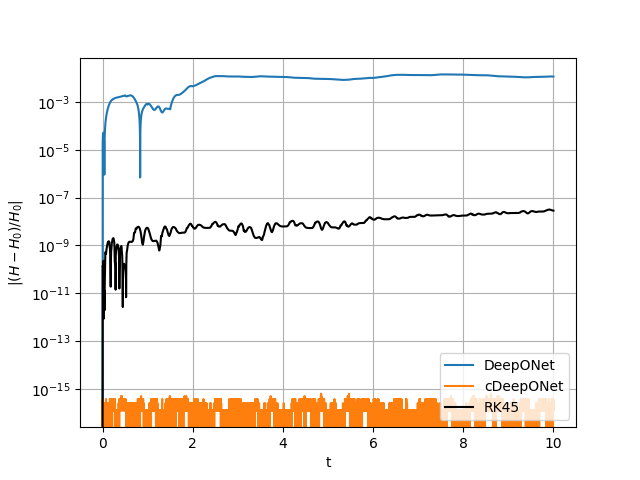}
\caption{Conservation of the Hamiltonian.}
\end{subfigure}
\caption{Numerical results for the double pendulum.}
\label{fig:DoublePendulumNumericalResults}
\end{figure}

For the sake of simplicity, we set $m_1=m_2=l_1=l_2=1$, and use $g=9.81$. We train vanilla and conservative DeepONets to solve the double pendulum problem. The initial conditions for $q_1$, $q_2$, $p_1$ and $p_2$ are sampled uniformly from $[-\pi/6,\pi/6]$ and each network is trained to obtain a solution for the time interval $t\in[0,0.5]$. We then use the trained DeepONets to time-step the solution for a total of 20 steps. The numerical results for one particular solution are presented in Figure~\ref{fig:DoublePendulumNumericalResults}.

These results show that neither of the numerical solutions is identical, which is consistent with the chaotic behaviour of the double pendulum. The relative error in the Hamiltonian verifies the exact conservation of the conservative DeepONet, whereas the standard DeepONet exhibits a systematic drift in the Hamiltonian.

\subsection{Conservative Lorenz--1963 model}

The Lorenz--1963 model~\cite{lore63Ay} reads
\begin{align}
\begin{split}
&\frac{\mathrm{d}x}{\mathrm{d}t} = \sigma(y - mx),\\
&\frac{\mathrm{d}y}{\mathrm{d}t} = x(\rho-z)-my,\\
&\frac{\mathrm{d}z}{\mathrm{d}t} = xy - \beta m z,
\end{split}
\end{align}
where $m\in\{0,1\}$ allows selecting between the standard ($m=1$) and conservative ($m=0$) Lorenz models. It was found in~\cite{nevi94Ay} that for the conservative Lorenz model, the following two first integrals exist
\[
H = z - \frac{x^2}{2\sigma},\quad L = \frac12(y^2+z^2) - \rho z.
\]

\begin{figure}[!ht]
\centering
\begin{subfigure}[b]{0.45\textwidth}
\includegraphics[width=\textwidth]{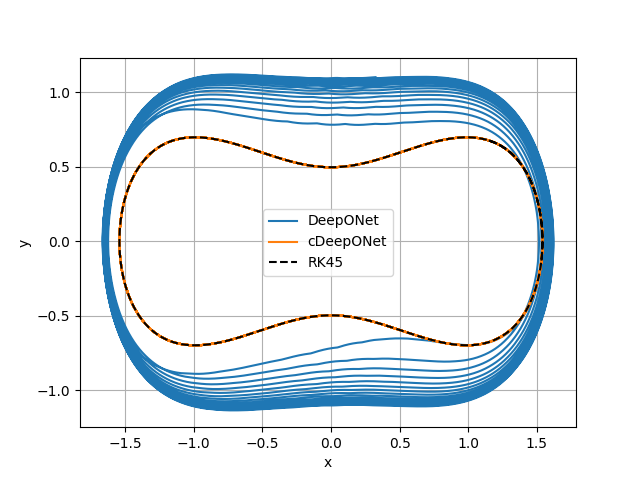}
\caption{Solution in the $(x,y)$ plane.}
\end{subfigure}
\begin{subfigure}[b]{0.45\textwidth}
\includegraphics[width=\textwidth]{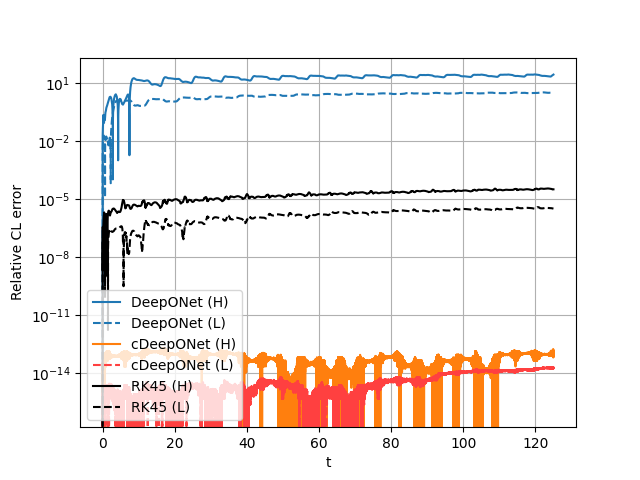}
\caption{Conservation of the first integrals.}
\end{subfigure}
\caption{Numerical results for the conservative Lorenz--1963 model.}
\label{fig:LorenzModelNumericalResults}
\end{figure}

In the sequel, we choose $\sigma=1$ and $\rho=0.5$ and train standard and conservative DeepONets. The initial conditions for $x_0$ and $y_0$ are sampled uniformly randomly from an annulus with inner radius of $0.5$ and outer radius of $1.5$ centered at the origin, and $z_0$ is sampled uniformly randomly from the interval $[0.5,1.5]$. The interval of integration was chosen as $t\in [0,0.25]$. After training we time-step the solution for 500 steps, with the numerical results for one particular solution being depicted in Figure~\ref{fig:LorenzModelNumericalResults}.

The solution for this problem should evolve along the intersection $\mathcal M$ of the two conserved surfaces defined by the implicit equations $H-H_0=0$ and $L-L_0=0$, and should therefore be closed and periodic. This is indeed the case for the conservative DeepONet, which exactly conserves both first integrals of the conservative Lorenz model. The standard DeepONet in turn again exhibits a spurious drift both in the first integrals and in the solution itself, and is not able to move along the intersection of the two conserved surfaces and instead moves outward to settle in an non-physical almost closed orbit.

\subsection{Point vortex problem}

As our final example, we consider the motion of $n$ point vortices in a plane~\cite{aref07a}. The equations of motion of this system are given by
\begin{align}
\begin{split}
    &\frac{\mathrm{d}x_i}{\mathrm{d}t} = -\frac{1}{2\pi}\sum_{j=0, j\ne i}^n\Gamma_j\frac{(y_i-y_j)}{r_{ij}^2},\\
    &\frac{\mathrm{d}y_i}{\mathrm{d}t} = \frac{1}{2\pi}\sum_{j=0, j\ne i}^n\Gamma_j\frac{(x_i-x_j)}{r_{ij}^2},
\end{split}
\end{align}
where $\{(x_i,y_i)\}_{i=0}^n$ are the positions of the $n$ vortices in the plane, $\Gamma_i\in\mathbb{R}$ is the vortex strength of the $i$th vortex, and $r_{ij}=\sqrt{(x_i-x_i)^2+(y_i-y_j)^2}$ is the Euclidean distance between vortices~$i$ and~$j$. This system possesses the following four first integrals
\[
H=-\frac{1}{4\pi}\sum_{1\leqslant i<j\leqslant n} \Gamma_i\Gamma_j\ln r_{ij},\quad  L =\sum_{i=1}^n\Gamma_i (x_i^2 + y_i^2),\quad P_x = \sum_{i=1}^n\Gamma_i x_i,\quad P_y=\sum_{i=1}^n\Gamma_i y_i,
\]
which are energy, angular momentum and linear momenta, respectively.

\begin{figure}[!ht]
\centering
\begin{subfigure}[b]{0.45\textwidth}
\includegraphics[width=\textwidth]{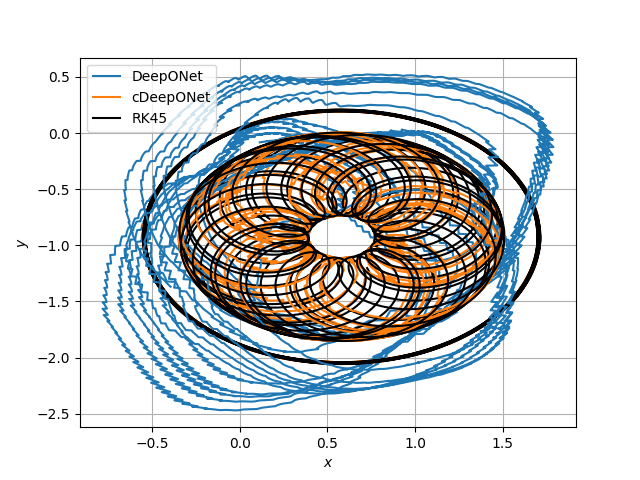}
\caption{Solution in the $(x,y)$ plane.}
\end{subfigure}
\begin{subfigure}[b]{0.45\textwidth}
\includegraphics[width=\textwidth]{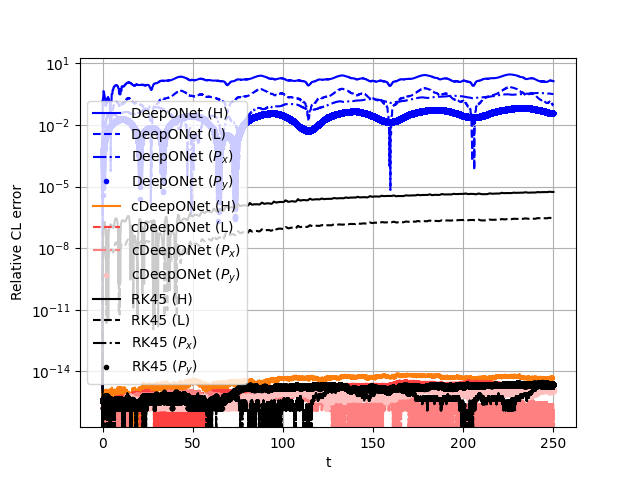}
\caption{Conservation of the first integrals.}
\end{subfigure}
\caption{Numerical results for a three vortex problem.}
\label{fig:PointVortexNumericalResults}
\end{figure}

We once again train both standard and conservative DeepONets for this problem, were we select the case of $n=3$ vortices. We set $\Gamma_i=1$, $i=1,2,3$, and sample the initial conditions for all vortices uniformly randomly from an annulus with inner radius of $0.5$ and outer radius of $1.5$ centered at the origin. The interval of integration was set to cover $t\in[0,0.5]$. After training both DeepONets, we use them for time-stepping particular initial conditions for a total of 500 steps with the numerical results being shown in Fig.~\ref{fig:PointVortexNumericalResults}.

This particular solution shows the interaction of two of the three vortices spiraling among their centre of vorticity, with the third vortex moving around them in a closed orbit with the same vortex centre. The standard DeepONet is not capable of capturing this behaviour, and does not preserve a single first integral of this problem. In turn, the solution of the conservative DeepONet by construction preserves all four first integrals with the vortices closely following the standard Runge--Kutta solution, which itself is only able to preserve the linear first integrals given by $P_x$ and $P_y$, cf.~\cite{hair06Ay}. 

\section{Conclusion}\label{sec:Conclusion}

We have introduced the notion of \textit{geometric numerical machine learning} and developed a method for exactly conservative physics-informed neural network solver for dynamical systems. The method relies on using a projection approach to map a learned candidate solution onto an invariant manifold spanned by the intersection of the iso-surfaces defined by the first integrals of the given problem. 

We have shown with several examples that the preservation of first integrals in dynamical systems is fundamental for the long-time integration of conservative systems, and greatly improves the qualitative and quantitative behaviour of the numerical solution obtained using conservative physics-informed neural network solver compared to their non-conservative (standard) counterparts. Notably, we have shown that standard DeepONets, which were introduced to allow for time-stepping of neural network based numerical integrators with the goal to overcome the shortcomings of physics-informed neural networks in obtaining solutions over longer time intervals, are not enough to give accurate long-time solutions to dynamical systems. For sufficiently long time intervals, even well-trained DeepONets lead to spurious behaviour that can be explained by a violation of the numerical conservation of first integrals. This situation parallels standard numerical methods, which are also not suitable for a long-time integration of conservative systems, an observation which gave rise to the rich field of geometric numerical integration~\cite{hair06Ay}.

By design, the proposed method works for all ordinary differential equations, and is thus applicable to a wide variety of problems of practical interest, including classical physics, molecular dynamics, mathematical biology and astronomy. The important case of partial differential equations, however, is not covered by the presented technique, which relies on the preservation of first integrals in each point along the trajectory of the solution of a dynamical system. In turn, conservation laws for partial differential equations require the preservation of conserved quantities over the entire spatial domain of the problem, which cannot be achieved with the point-wise projection approach suitable for dynamical systems. Therefore, different techniques will have to be developed for conservative physics-informed neural networks and conservative DeepONets for partial differential equations.

\begin{ack}
This research was undertaken, in part, thanks to funding from the Canada Research Chairs program and the NSERC Discovery Grant program. The authors thank Seth Taylor for helpful discussions.
\end{ack}

{\footnotesize\setlength{\itemsep}{0ex}


\begin{thebibliography}{10}

\bibitem{aref07a}
H.~Aref.
\newblock Point vortex dynamics: a classical mathematics playground.
\newblock {\em J. Math. Phys.}, 48(6), 2007.

\bibitem{aror23a}
S.~Arora, A.~Bihlo, and F.~Valiquette.
\newblock Invariant physics-informed neural networks for ordinary differential
  equations.
\newblock {\em arXiv preprint arXiv:2310.17053}, 2023.

\bibitem{bihl23a}
A.~Bihlo.
\newblock Improving physics-informed neural networks with meta-learned
  optimization.
\newblock {\em arXiv preprint arXiv:2303.07127}, 2023.

\bibitem{bihl22a}
A.~Bihlo and R.~O. Popovych.
\newblock Physics-informed neural networks for the shallow-water equations on
  the sphere.
\newblock {\em J. of Comput. Phys.}, 456:111024, 2022.

\bibitem{bihl17b}
A.~Bihlo and F.~Valiquette.
\newblock Symmetry-preserving numerical schemes.
\newblock In {\em Symmetries and Integrability of Difference Equations}, pages
  261--324. Springer, 2017.

\bibitem{brec23b}
R.~Brecht and A.~Bihlo.
\newblock {M-ENIAC}: {A} machine learning recreation of the first successful
  numerical weather forecasts.
\newblock {\em arXiv preprint arXiv:2304.09070}, 2023.

\bibitem{brec23c}
R.~Brecht, D.~R. Popovych, A.~Bihlo, and R.~O. Popovych.
\newblock Improving physics-informed {DeepONets} with hard constraints.
\newblock {\em arXiv preprint arXiv:2309.07899}, 2023.

\bibitem{brow20a}
T.~Brown, B.~Mann, N.~Ryder, M.~Subbiah, et~al.
\newblock Language models are few-shot learners.
\newblock In H.~Larochelle, M.~Ranzato, R.~Hadsell, M.F. Balcan, and H.~Lin,
  editors, {\em Advances in Neural Information Processing Systems}, volume~33,
  pages 1877--1901. Curran Associates, Inc., 2020.

\bibitem{chen95a}
T.~Chen and H.~Chen.
\newblock Universal approximation to nonlinear operators by neural networks
  with arbitrary activation functions and its application to dynamical systems.
\newblock {\em IEEE Trans. Neural Netw.}, 6(4):911--917, 1995.

\bibitem{cuom22a}
S.~Cuomo, V.~S. Di~Cola, F.~Giampaolo, G.~Rozza, M.~Raissi, and F.~Piccialli.
\newblock Scientific machine learning through physics--informed neural
  networks: where we are and what's next.
\newblock {\em J. Sci. Comput.}, 92(3):88, 2022.

\bibitem{finz21a}
M.~Finzi, M.~Welling, and A.~G. Wilson.
\newblock A practical method for constructing equivariant multilayer
  perceptrons for arbitrary matrix groups.
\newblock In {\em ICML}, pages 3318--3328. PMLR, 2021.

\bibitem{gold80Ay}
H.~Goldstein.
\newblock {\em Classical {M}echanics}.
\newblock Addison-Wesley, Reading, 1980.

\bibitem{hair06Ay}
E.~Hairer, C.~Lubich, and G.~Wanner.
\newblock {\em {Geometric numerical integration: structure-preserving
  algorithms for ordinary differential equations}}.
\newblock Springer, Berlin, 2006.

\bibitem{holm09a}
D.~D. Holm, T.~Schmah, and C.~Stoica.
\newblock {\em Geometric mechanics and symmetry: from finite to infinite
  dimensions}, volume~12.
\newblock Oxford University Press, 2009.

\bibitem{jagt20a}
A.~D. Jagtap, E.~Kharazmi, and G.~E. Karniadakis.
\newblock Conservative physics-informed neural networks on discrete domains for
  conservation laws: applications to forward and inverse problems.
\newblock {\em Comput. Methods Appl. Mech. Eng.}, 365:113028, 2020.

\bibitem{kris21a}
A.~Krishnapriyan, A.~Gholami, S.~Zhe, R.~Kirby, and M.~W. Mahoney.
\newblock Characterizing possible failure modes in physics-informed neural
  networks.
\newblock {\em Advances in Neural Information Processing Systems},
  34:26548--26560, 2021.

\bibitem{kriz12a}
A.~Krizhevsky, I.~Sutskever, and G.~E. Hinton.
\newblock Imagenet classification with deep convolutional neural networks.
\newblock In {\em Advances in Neural Information Processing Systems},
  volume~25, pages 1097--1105. Curran Associates, 2012.

\bibitem{laga98a}
I.~E. Lagaris, A.~Likas, and D.~I. Fotiadis.
\newblock Artificial neural networks for solving ordinary and partial
  differential equations.
\newblock {\em IEEE Trans. Neural Netw.}, 9(5):987--1000, 1998.

\bibitem{levi93a}
R.~B. Levien and S.~M. Tan.
\newblock Double pendulum: {A}n experiment in chaos.
\newblock {\em Am. J. Phys.}, 61(11):1038--1044, 1993.

\bibitem{lore63Ay}
E.~N. Lorenz.
\newblock Deterministic nonperiodic flow.
\newblock {\em J.~\mbox{Atmos}. Sci.}, 20(2):130--141, 1963.

\bibitem{lu21a}
L.~Lu, P.~Jin, G.~Pang, Z.~Zhang, and G.~E. Karniadakis.
\newblock Learning nonlinear operators via {DeepONet} based on the universal
  approximation theorem of operators.
\newblock {\em Nat. Mach. Intell.}, 3(3):218--229, 2021.

\bibitem{mulle23a}
E.~H. M{\"u}ller.
\newblock Exact conservation laws for neural network integrators of dynamical
  systems.
\newblock {\em J. Comput. Phys.}, 488:112234, 2023.

\bibitem{namb73Ay}
Y.~Nambu.
\newblock Generalized {H}amiltonian dynamics.
\newblock {\em Phys. Rev. D}, 7(8):2405--2412, 1973.

\bibitem{nevi94Ay}
P.~N\'{e}vir and R.~Blender.
\newblock Hamiltonian and {N}ambu representation of the non-dissipative
  {L}orenz equations.
\newblock {\em Beitr. Phys. Atmosph.}, 67(2):133--140, 1994.

\bibitem{rais18a}
M.~Raissi.
\newblock Deep hidden physics models: Deep learning of nonlinear partial
  differential equations.
\newblock {\em J.~Mach. Learn. Res.}, 19(1):932--955, 2018.

\bibitem{rais19a}
M.~Raissi, P.~Perdikaris, and G.~E. Karniadakis.
\newblock Physics-informed neural networks: A deep learning framework for
  solving forward and inverse problems involving nonlinear partial differential
  equations.
\newblock {\em J. Comput. Phys.}, 378:686--707, 2019.

\bibitem{sanz94a}
J.-M. Sanz-Serna and M.-P. Calvo.
\newblock {\em Numerical {H}amiltonian problems}, volume~7 of {\em Applied
  Mathematics and Mathematical Computation}.
\newblock Chapman \& Hall, London, 1994.

\bibitem{silv17a}
David Silver, Julian Schrittwieser, Karen Simonyan, Ioannis Antonoglou, Aja
  Huang, Arthur Guez, Thomas Hubert, Lucas Baker, Matthew Lai, Adrian Bolton,
  et~al.
\newblock Mastering the game of {G}o without human knowledge.
\newblock {\em Nature}, 550(7676):354--359, 2017.

\bibitem{wan16a}
A.~T.~S. Wan, A.~Bihlo, and J.-C. Nave.
\newblock Conservative methods for dynamical systems.
\newblock {\em SIAM J. Numer. Anal.}, 55(5):2255--2285, 2017.

\bibitem{wang23a}
S.~Wang and P.~Perdikaris.
\newblock Long-time integration of parametric evolution equations with
  physics-informed {DeepONets}.
\newblock {\em J. Comput. Phys.}, 475:111855, 2023.

\bibitem{wang22a}
S.~Wang, S.~Sankaran, and P.~Perdikaris.
\newblock Respecting causality is all you need for training physics-informed
  neural networks.
\newblock {\em arXiv preprint arXiv:2203.07404}, 2022.

\end{thebibliography}
\end{document}